\documentclass[review]{elsarticle}

\usepackage{hyperref}
\usepackage{amssymb}

\journal{Journal of \LaTeX\ Templates}

\begin{document}

\begin{frontmatter}

\title{C-RPNs: Promoting Object Detection in real world via a Cascade Structure of Region Proposal Networks}

\author[mymainaddress,myfootnote]{Dongming Yang}
\ead{yangdongming@pku.edu.cn}
\author[mymainaddress,mysecondaryaddress]{YueXian Zou\corref{mycorrespondingauthor}}
\ead{zouyx@pku.edu.cn}
\ead[url]{http://web.pkusz.edu.cn/adsp/}
\fntext[myfootnote]{Address: Peking University, Xili Road, Nanshan District, Shenzhen, P.R.China}
\author[mythirdaddress]{Jian Zhang}
\ead{Jian.Zhang@uts.edu.au}
\author[mymainaddress,mysecondaryaddress,myforthaddress]{Ge Li}
\ead{gli@pkusz.edu.cn}
\cortext[mycorrespondingauthor]{Corresponding author}

\address[mymainaddress]{ADSPLAB, School of ECE, Peking University, Shenzhen, China}
\address[mysecondaryaddress]{Peng Cheng Laboratory, Shenzhen, China}
\address[mythirdaddress]{School of Electrical and Data Engineering, University of Technology Sydney, Australia}
\address[myforthaddress]{National Engineering Laboratory for Video Technology - Shenzhen Division}

\begin{abstract}
Recently, significant progresses have been made in object detection on common benchmarks (i.e., Pascal VOC). However, object detection in real world is still challenging due to the serious data imbalance. Images in real world are dominated by easy samples like the wide range of background and some easily recognizable objects, for example. Although two-stage detectors like Faster R-CNN achieved big successes in object detection due to the strategy of extracting region proposals by region proposal network, they show their poor adaption in real-world object detection as a result of without considering mining hard samples during extracting region proposals. To address this issue, we propose a Cascade framework of Region Proposal Networks, referred to as C-RPNs. The essence of C-RPNs is adopting multiple stages to mine hard samples while extracting region proposals and learn stronger classifiers. Meanwhile, a feature chain and a score chain are proposed to help learning more discriminative representations for proposals. Moreover, a loss function of cascade stages is designed to train cascade classifiers through backpropagation. Our proposed method has been evaluated on Pascal VOC and several challenging datasets like BSBDV 2017, CityPersons, etc. Our method achieves competitive results compared with the current state-of-the-arts and all-sided improvements in error analysis, validating its efficacy for detection in real world.
\end{abstract}

\begin{keyword}
\texttt Object Detection\sep Hard Samples Mining \sep Cascade Network \sep Region Proposal Network
\end{keyword}

\end{frontmatter}


\section{Introduction}
Object detection is a most fundamental step in visual understanding, which aims at identifying and localizing objects of certain categories in images. To promote the development of object detection, plenty of benchmarks have been developed, i.e., PASCAL VOC \cite{everingham2010the} and MS COCO \cite{lin2014microsoft}. Most of object detection approaches are trained and tested on these common object detection benchmarks, which typically assume that objects in images are with good visibility and balance. Obviously, this assumption is usually not satisfied in real world.

\begin{figure}
	\centering
		\includegraphics[scale=.28]{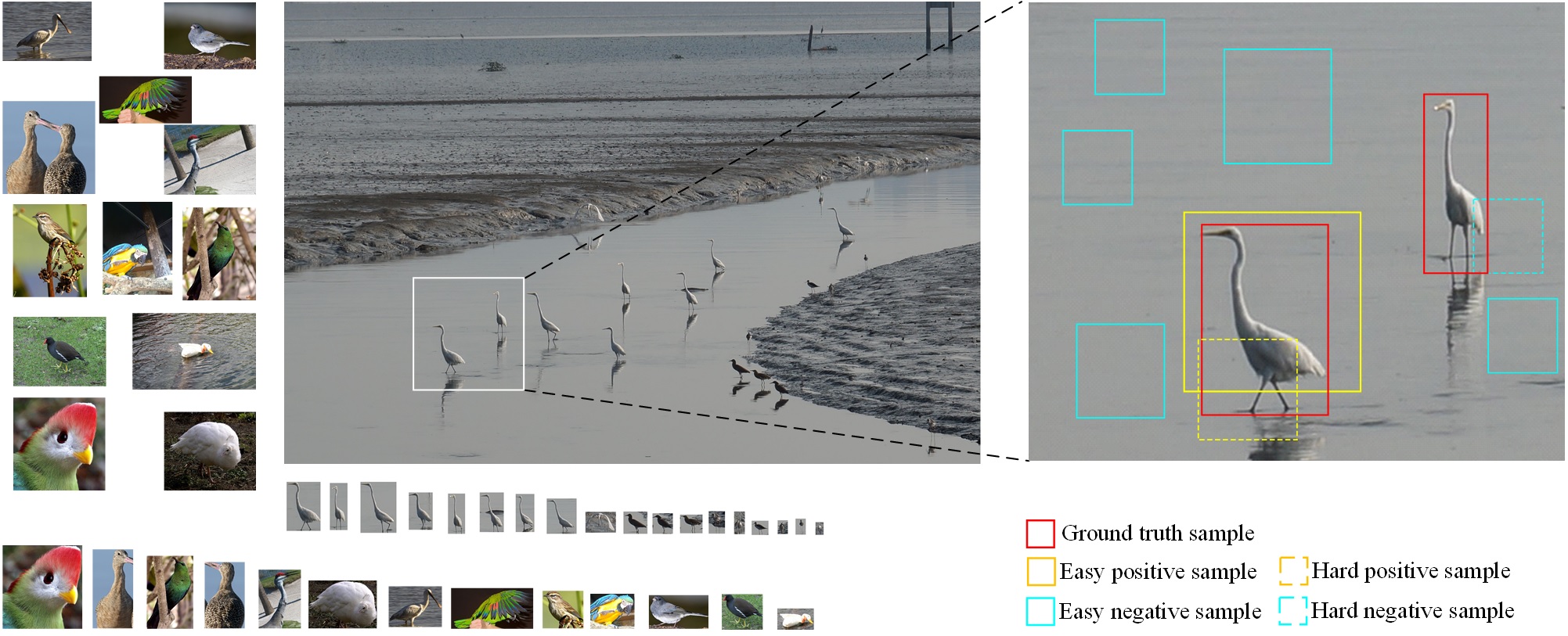}
	\caption{Examples: (1) 12 images from Pascal VOC (left upper); (2) one littoral bird image from BSBDV 2017 (middle upper); (3) The bird objects drawn from these images (bottom); (4) The easy and hard samples listed randomly from the real-world image (right). It is clearly that the image from realistic scenes is dominated by easy samples, especially easy negative samples. Besides, the scales and abundances of birds are mismatched from common benchmarks to realistic scenes, Best viewed in color.}
	\label{FIG:1}
\end{figure}

Taking littoral bird images from developed benchmarks and wild scenes as examples, the former are usually collected with better visibility, while the latter are collected via monitoring cameras with different background and camera distance. Moreover, different illumination and weather conditions may appear in wild scenes. For more intuitive observation, several examples of littoral birds are illustrated in  Figure~\ref{FIG:1}. The image from BSBDV 2017 \cite{guan2018multi-scale} shows birds from wild scenes, while images from PASCAL VOC \cite{everingham2010the} show birds from common benchmarks. The image from BSBDV 2017 is with resolution of 4912*3264, in which the heights of birds vary from 80 to 300 pixels. Images from PASCAL VOC 2007 and 2012 are with average resolution of 400*400, where the heights of birds are from 150 to 480 pixels. Apparently, the easily recognizable background in wild scenes take more prominent position compared with that in common benchmarks. Besides, bird objects obtained from the wild scenes are with smaller sizes and less texture information. For object detection techniques, such a distribution mismatch from common benchmarks to real world have been observed to lead to a significant performance degradation.

Although enriching training data could possibly alleviate the performance degradation, it is not favored since annotating data is expensive and time consuming. Therefore, developing object detectors towards real world is desirable. To figure out the crucial elements of performance degradation in real-world object detection, plenty of experiments have been conducted. We list the conclusions as follows:
\begin{enumerate}[(1)]
\item Data imbalance frequently occurs in real world. From an image in real world, the number of negative samples (also called background samples) is much larger than that of positive samples (As shown in Figure~\ref{FIG:1}), and most of them are easy samples. Easy samples do not contribute useful learning information during training while hard samples benefit the convergence and the detection accuracy. Thus, the overwhelming number of easy samples during training leads to moronic classifiers and degenerate models.
\item As mentioned above, because of the smaller size, poor shooting conditions and poor abundance of objects in real-world scenes, classifiers in detection algorithms are unable to learn discriminative features from ground truth.
\end{enumerate}

In this work, we aim to improve the precision of object detection in real world. Based on observations above, mining hard samples from abundant easy samples for training is a crucial route to address this issue. Based on the brilliant object detector Faster R-CNN \cite{ren2017faster}, we firstly propose a cascade framework of region proposal networks, referred to as C-RPNs. While extracting region proposals, C-RPNs are adopted to mine hard samples and learn stronger classifiers. Multi-stage classifiers at early stages discard most of easy samples so that classifiers at latter stages focus on handling hard samples. Also, we design a feature chain and a score chain to generate more discriminative representations for proposals. Finally, a loss function of cascade stages is built to jointly learn cascade classifiers.

The contributions of this work are summarized as follows:
\begin{itemize}
\item Based on the Faster R-CNN, a cascade structure of region proposal networks for object detection was firstly proposed, referred to as C-RPNs.
\item A feature chain and a score chain were designed in C-RPNs to further improve the classification capacity of multi-stage classifiers.
\item A loss function of multi-stage was constructed to jointly learn cascade classifiers.
\item Integrating the proposed components into the Faster R-CNN, our resulting model can be trained end-to-end.
\end{itemize}
Extensive experiments have been conducted on several datasets, including PASCAL VOC \cite{everingham2010the} , BSBDV 2017 \cite{guan2018multi-scale}, Caltech Pedestrian Benchmark \cite{dollar2009pedestrian} and CityPersons \cite{zhang2017citypersons:}. Our approach have provided competitive performance compared with the current state-of-the-arts. Besides, error analyses have shown that our approach achieved all-sided improvements compared with the baseline Faster R-CNN. The experimental results demonstrate the effectiveness of our proposed approach for object detection in real world.

\section{Related work}

\subsection{Related Work On Object Detection}

We all have witnessed tremendous progresses in object detection using convolutional neural networks (CNNs) in recent years \cite{girshick2014rich,girshick2015fast,ren2017faster,dai2016r-fcn:,gidaris2015object,liu2016ssd:,redmon2016you}. Region-based CNN approaches \cite{girshick2014rich,girshick2015fast,ren2017faster} are referred as two-stage detectors, which have received great attention due to their effectiveness. At the outset, R-CNN \cite{girshick2014rich} was constrained by a selected search region. To reduce the computational complexity of R-CNN, Fast R-CNN \cite{girshick2015fast} shared the convolutional feature maps among region of interest (RoI) and accelerated spatial pyramid pooling using RoI pooling layer. Renetal \cite{ren2017faster} introduced Region Proposal Network (RPN) to generate high-quality region proposals and then merged them with Fast R-CNN into a single network, referred to as Faster R-CNN. Besides, for faster detection, one-stage detectors such as YOLO \cite{redmon2016you} and SSD \cite{liu2016ssd:} were proposed to accomplish detection without region proposals, although this strategy reduced the detection performance. Researches showed that Faster R-CNN achieved a big success in object detection and laid the foundation for many follow-up works \cite{gidaris2015object,lin2017feature,zhang2016is,dai2017deformable}. For example, feature pyramid and fusion operations were adopted \cite{lin2017feature} to enhanced precision of detection. Deeper \cite{simonyan2015very,he2016deep,szegedy2015going} or wider \cite{bell2016inside-outside,zagoruyko2016wide} networks also benefited the performance. Deformable CNN \cite{dai2017deformable} and Receptive Field Block Net \cite{liu2018receptive} enhanced the convolutional features using deformable convolutional operation and Receptive Field Block respectively. In addition, using large batch size \cite{peng2018megdet:} during training provided improvement in detection. SIN \cite{liu2018structure} jointly used scene context and object relationships to promote detection performance.

Although reasonable detection performances have been achieved on benchmarks like PASCAL VOC \cite{everingham2010the} and MS COCO \cite{lin2014microsoft}, object detection in real world still suffers from poor precision. Works mentioned above mostly focused on the conventional setting while rarely considered the adaptation issues for object detection in real world such as data imbalance.

\subsection{Related Work On Hard Example Mining and Cascade CNN}

Gradually updating the set of background samples by selecting those from samples which are detected as false positives, bootstrapping \cite{sung1996learning} was the earliest solution to automatic employ hard samples for training. The strategy in bootstrapping led to an iterative process that alternates between updating the trained model and finding new false positives to add to the bootstrapped training set. Bootstrapping techniques were then successfully applied on detectors driven by CNN and SVMs for object detection \cite{girshick2014rich,he2014spatial}, generally referred to as hard negative mining. After that, CNN detectors like Fast R-CNN \cite{girshick2015fast} and its descendants were trained with stochastic gradient descent (SGD) on millions of samples, in which bootstrapping as an offline progress was no longer been adopted. To balance positive and negative training samples but without thinking of mining hard ones, Faster R-CNN \cite{ren2017faster} randomly used 256 samples in an image to compute the loss function of a mini-batch, where the positive and negative ones have a ratio of up to 1:1. A number of methods \cite{simoserra2014fracking,loshchilov2015online,shrivastava2016training} then focused on mining hard samples online for training convolutional networks. Rowley \cite{simoserra2014fracking} selected hard positive and negative samples from a larger set of random samples based on their loss independently. Sermanet\cite{loshchilov2015online} and Shrivastava \cite{shrivastava2016training} focused on online hard sample selection strategies for mini-batch SGD methods and then OHEM \cite{shrivastava2016training} were introduced for region-based detectors which built mini-batches with the highest-loss samples. Recently, Focal Loss \cite{lin2017focal} has been proposed to address the extreme foreground-background class imbalance problem in object detection with one-stage detectors, which applied a modulating term to the cross entropy loss in order to focus learning on hard negative examples. Analyzing of previous works shows that inchoate bootstrapping techniques are inappropriate for CNN-based detectors. Some online hard example mining strategies selected hard examples based on their loss, which are innovative but time-consuming. Focal Loss focused on dealing with data imbalance with one-stage detectors, while our works pay more attention to two-stage detectors with region proposals.

From another perspective, cascade structure is a widely used technique to discard easy samples at early stages for learning better classification models. Before the prosperity of CNNs, cascade structure were applied to SVM \cite{felzenszwalb2010cascade} and boosted classifiers \cite{dollar2014fast,xiao2003boosting} with hand-crafted features. Multi-stage classifiers have been proved to be effective in generic object detection \cite{felzenszwalb2010cascade} and face detection \cite{xiao2003boosting,bourdev2005robust}, although these multiple classifiers were not trained jointly. It was showed that CNNs with cascade structure performed effectively on classification \cite{ouyang2015deepid-net:,li2015a,yang2016craft} as well, in which multiple but separate CNNs were trained. After that, Qin \cite{qin2016joint} proposed a method to jointly train a cascade CNNs. The recent method Cascade R-CNN \cite{cai2018cascade} trained Faster R-CNN with cascade increasing IoU thresholds, which was innovative, but without considering the data imbalance issue.
Based on observations above, cascade structures are potential, but existing works either cannot be aggregated in the R-CNN based detection framework or have not considered building cascade structure on RPN to help extracting hard region proposals. Thus, confronting with the data imbalance problem in object detection in real world, the existing outcomes are very limited.

In this work, we propose C-RPNs to mine hard samples while extracting region proposals and learn more discriminative features for object detection in real world. Integrating with Faster R-CNN model, our proposed method, to the best of our knowledge, is the first cascade model of region proposal networks for object detection.

\section{Proposed Method}

\subsection{Overview Of C-RPNs}

Faster R-CNN consists of a shared backbone convolutional network, a region proposal network (RPN) and a final classifier based on region-of-interest (RoI), in which the RPN is employed to extract region proposals. Without considering mining hard samples in the process of RPN, Faster R-CNN shows its limited capacity in detection in realistic scenes.  Our novel C-RPNs are firstly proposed to address this problem.

\begin{figure}
	\centering
		\includegraphics[scale=.42]{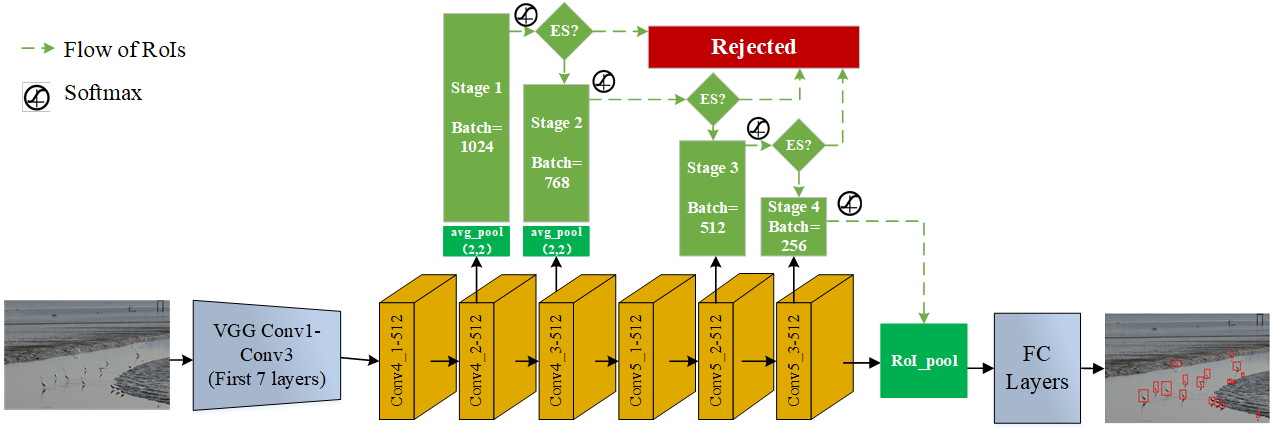}
	\caption{An overall of our proposed C-RPNs model. We adopt VGG16 as backbone network. ES refers to easy samples.}
	\label{FIG:2}
\end{figure}

For performance comparison fairness, VGG16 is taken as the backbone network \cite{simonyan2015very}. Figure~\ref{FIG:2} shows an overview of our proposed C-RPNs model. At first, several shared bottom convolutional layers are used for extracting convolutional features from the image (Conv1-Conv4\_1). Then, C-RPNs is adopted upon four different convolutional layers, which are Conv4\_2, Con4\_3, Conv5\_2 and Conv5\_3. Since feature maps from Conv5 have the same channels but half size compared with those from Conv4, we employ an average pooling with size of 2*2 upon Conv4\_2 and Conv4\_3 to obtain feature maps of same resolutions for these four stages.
At stage \emph{1}, the feature map extracted from Conv4\_2 are used for generating region proposals and obtaining binary classification scores by a softmax function. The binary classification scores estimate a sample’s probabilities belonging to background and objects. With the classification scores and a reject threshold \emph{r}, part of easy samples will be rejected at this stage, which are detailed in Section 3.3. At stage \emph{2}, if a proposal has not been rejected at the former stage, then the feature map from Conv4\_3 for this proposal is used for further binary classification. Similar processes are applied at stage \emph{3} and stage \emph{4}. Since there is no constrain that the rejected samples must be background, few easy positive samples might also be rejected at early stages during training. It is worth to point out that the stage \emph{4} achieves not only binary classification but also bounding box regression. After these four stages, the proposals have not been rejected are sent to RoI pooling layer for final detection. In this study, we set batch of each stage as 1024, 768, 512 and 256 respectively so that the stage \emph{4} has the same batch size with RPN from Faster R-CNN. It is worth mentioning that the reason why we set only \emph{4} stages not \emph{5} or more is that employing the shallow and bigger feature maps from Conv3 contributes very limited performance gain but is time-consuming according to our experiments.

From Figure~\ref{FIG:2}, it can be seen that C-RPNs takes different convolutional features stage-by-stage which enables it obtains different semantic information and receptive field. It is also noted that, in C-RPNs, the classifiers at shallow stages handle easier samples so that the classifiers at deeper stages focus on handling more difficult samples. The easy samples rejected by a classifier from shallow stage will not participate in the latter stages. With this design, abundant samples can be used but only hard samples been mined will go for final classification and bounding box regression, which benefits to alleviate the data imbalance problem.

To further enhance the classification capacity, a feature chain and a score chain are designed in C-RPNs, which are detailed in Section 3.2. In the end, the multi-stage classifications and bounding box regressions are learned in an end-to-end manner through backpropagation via a joint loss function, details are given in Section 3.3.

\subsection{Feature Chain and Score Chain}

\begin{figure}
	\centering
		\includegraphics[scale=.6]{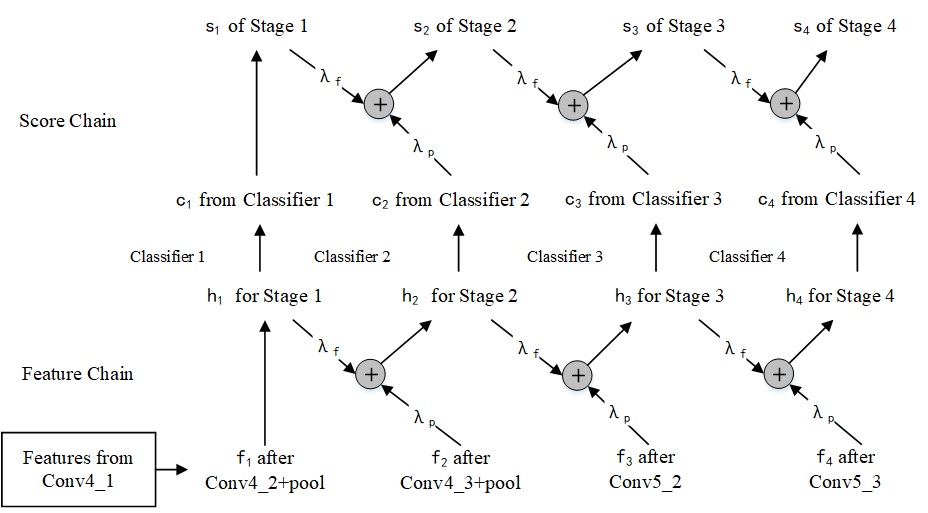}
	\caption{The proposed feature chain and score chain of C-RPNs.}
	\label{FIG:3}
\end{figure}

Literature studies show that FPN \cite{lin2017feature} and DSSD \cite{fu2017dssd} are effective for object detection using multiple convolutional layers. In this study, in order to capture the variation of features from different layers, a feature chain and a score chain at cascade stages are designed which are able to make use of features at previous stages as the prior knowledge for the classification at current stage. Not like the top-down pathway and lateral connections from FPN, our feature fusion operation follows the bottom-up pathway, which is the feed-forward computation of the VGG16. The description of feature chain and score chain is shown in Figure~\ref{FIG:3}.

We define the number of stages as \emph{T} and \emph{t} is the stage index. At stage \emph{t}, we denote the features from convolutional layer as \emph{$f_t$} while features for classification as \emph{$h_t$}. The feature chain is formulated as following:
\begin{equation}
h_t=\left\{\begin{array}{ccc}
f_t & when & t=1 \\
\lambda_f*h_{t-1}\bigoplus\lambda_p*f_t & when & t > 1
\end{array}\right.
\label{equ1}
\end{equation}
where $\oplus$ denotes the summarized point to point. \emph{$\lambda=\{\lambda_f,\lambda_p\}$} are hyper parameters controlling the weight of features from former stage and present convolutional layer to generate fusional features for classification. \emph{$\lambda_f$} and \emph{$\lambda_p$} add up to 1. Considering features from present convolutional layer are more helpful for classification, we set \emph{$\lambda_f$} as 0.1 and \emph{$\lambda_p$} as 0.9 according to our empirical tests (detailed in Section 4.5). The fused features \emph{$h_t$} are then used for classification.

At stage \emph{t}, for each proposal have not been rejected at the \emph{t-1} stage, we denote the score from classifier t as \emph{$c_t$} while the output score of this stage as \emph{$s_t$}. The designed score chain has the following formulation.
\begin{equation}
s_t=\left\{\begin{array}{ccc}
c_t & when & t=1 \\
\lambda_f*s_{t-1}+\lambda_p*c_t & when & t > 1
\end{array}\right.
\label{equ1}
\end{equation}
In this implementation, features and scores at current stage make use of those from previous stages which enhance the capacity of the classifiers at current stage.

\subsection{Cascade Loss Function with Samples Mining}

\begin{figure}
	\centering
		\includegraphics[scale=.6]{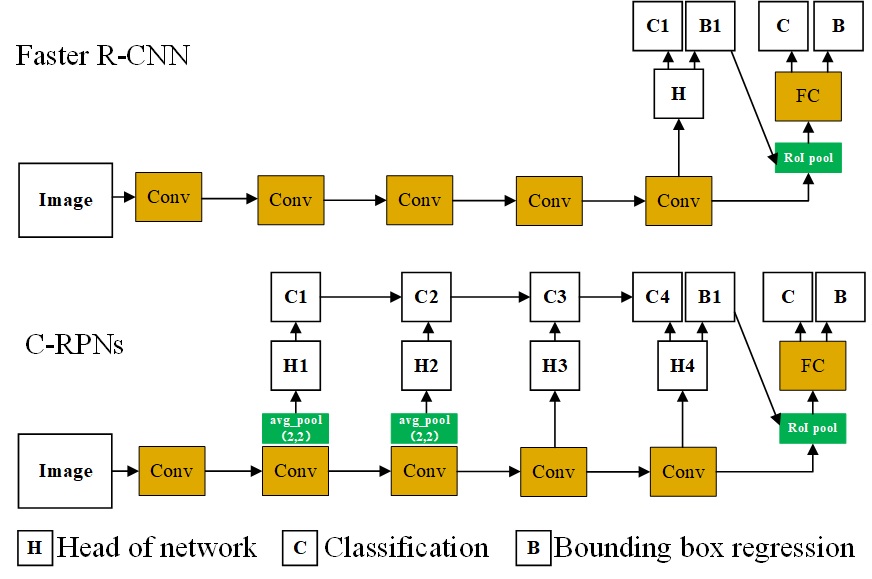}
	\caption{Cascade Losses of our proposed C-RPNs. Faster R-CNN \cite{ren2017faster} is displayed as baseline to show our characteristics.}
	\label{FIG:4}
\end{figure}

In Faster R-CNN, training loss is composed of loss from RPN and ROIs. The former contains a binary classification loss and a regression loss. In our method, illustrated in Figure~\ref{FIG:4}, C-RPNs contains four binary classification losses and a regression loss.

In C-RPNs, the cascade classifiers assign a sample’s probabilities to background and objects. \emph{k=\{0,1\}} is denoted to express these two class respectively. At stages \emph{$t\epsilon\{1,2,3,4\}$}, the set of class scores for a sample are denoted by \emph{$s=\{s_t | t=1,\ldots,T\}$}. \emph{$s_t=\{s_{(t,0)},s_{(t,1)}\}$} are scores at stage \emph{t} for background and objects respectively. Another layer at stage \emph{4} outputs bounding box regression offsets  \emph{$l=\{l^k | k=1\}$}, \emph{$l^k=(l_x^k,l_y^k,l_w^k,l_h^k)$} for objects. Our proposed loss function of C-RPNs has the following formulation:
\begin{equation}
\L_{C-RPNs}(s,k^*,l,l^*)=L_{cls}(s,k^*)+L_{loc}(l,l^*,k^*)\\
\label{equ3}
\end{equation}
\begin{equation}
\L_{cls}(s,k^*)=-\sum_{t=1}^{T}\alpha_{t}\mu_{t}log(s_{t,k^*})\\
\label{equ4}
\end{equation}
where \emph{$L_{cls} (*)$} is the loss for classification and \emph{$L_{loc}$} is the loss for bounding box regression. For \emph{$L_{loc}$}, we use the \emph{$smoothed$ $L_1$} loss \cite{girshick2015fast}. For \emph{$L_{cls} (*)$}, \emph{$\alpha_t$} and \emph{$\mu_t$} are defined as follows, where \emph{$\alpha_t$} is a parameter that controls the weight of loss from cascade classifiers and \emph{$\mu_t$} evaluates whether the sample is rejected at previous stages.
\begin{equation}
\alpha_{t}=\frac{\alpha_{T}}{10^{T-t}}
\label{equ5}
\end{equation}
\begin{equation}
\mu_t=\left\{\begin{array}{ccc}
\prod_{i=1}^{t-1}[s_{t,k^*}<r] & when & t>1  \\
1 & when & t= 1
\end{array}\right.
\label{equ6}
\end{equation}
Here, we set \emph{$\alpha_T=1$}, where \emph{T=4} in C-RPNs. Since scores from deeper classifiers are more crucial for final classification than those from shallow classifiers, \emph{$\alpha_t$} from deeper classifiers has been distributed more weight with a tenfold increase based on our experience. For \emph{$\mu_t$}, we set the  \emph{r} as a threshold value at each stage. \emph{$[s_{(t,k^* )}<r]$} will output  \emph{1} if it is true or output  \emph{0} if it is false. If a sample has been rejected at previous stages, it will no longer be used for training the classifier at current stage. We set \emph{r} as \emph{0.99} according to our empirical tests (detailed in Section 4.5). If \emph{$\alpha_t=\mu_t=1$} and \emph{$T=1$}, then \emph{$L_{cls} (*)$} is a normal cross entropy loss.

For the object detection with the proposed model, the final training loss is designed to compose the loss from C-RPNs and the loss from ROIs:
\begin{equation}
L_{detection}=L_{C-RPNs}+L_{roi}
\label{equ7}
\end{equation}
where \emph{$L_{C-RPNs}$} and \emph{$L_{roi}$} both are composed of classification loss and regression loss. The former contains four cascade binary classification losses while the latter contains a multi-class classification loss. With this loss function, multiple classifiers and bounding box regressions are learned jointly through backpropagation.

\section{Experiments and Evaluations}

\subsection{Experimental setup}

\paragraph{ Datasets and Evaluation Metrics} We evaluated our approach on several public object detection datasets, including PASCAL VOC \cite{everingham2010the} , BSBDV 2017 \cite{guan2018multi-scale}, Caltech Pedestrian Benchmark \cite{dollar2009pedestrian} and CityPersons \cite{zhang2017citypersons:}. For evaluation, we used the standard average precision (AP) and mean average precision (mAP) scores with IoU thresholds at \emph{0.5}.

Pascal VOC. Pascal VOC involves 20 categories. VOC 2007 dataset consists of about 5k trainval images and 5k test images, while VOC 2012 dataset includes about 11k trainval images and 11k test images. Following the protocol in \cite{girshick2015fast}, we perform training on the union of VOC 2007 trainval and VOC 2012 trainval. The test is conducted on VOC 2007 test set.

BSBDV2017. The Birds Dataset of Shenzhen Bay in Distant View \cite{guan2018multi-scale} is a great challenging dataset in wild scenes, consisting of 1,421 trainval images and 351 test images. BSBDV2017 contains three kinds of image resolutions, which are 2736*1824, 4288*2848 and 5472*3648 respectively. Size of birds varies greatly from 18*30 to 1274*632.

Caltech Pedestrian Benchmark. The Caltech Pedestrian Benchmark \cite{dollar2009pedestrian} includes a total of 350,000 bounding boxes of pedestrians. Approximately 2,300 unique pedestrians were annotated in roughly 250,000 frames. Following the protocol in \cite{yang2018real-time}, one frame from every five frames of Caltech Benchmark and all frames of the ETH \cite{ess2008a} and TUD-Brussels \cite{wojek2009multi-cue} are extracted as training data, which includes 27,021 images in total. 4,024 images in the standard test set are used for evaluation.

CityPersons. The CityPersons \cite{zhang2017citypersons:} consists of images recorded across 27 cities, 3 seasons, various weather conditions and more common crowds. It creates high quality bounding box annotations for pedestrians in 5000 images, which is a subset of the Cityscapes dataset \cite{cordts2016the}. 2975 images from train set and 500 images from val set are used for training and testing respectively.

\paragraph{Implementation Details}Faster R-CNN is taken as our baseline, where all parameters are set according to the original publication \cite{ren2017faster} if not specified. We initialize the backbone network using a VGG16 pre-trained model on ImageNet \cite{deng2009imagenet:} while all new layers are initialized by drawing weights from a zero-mean Gaussian distribution with standard deviation 0.01. For training on Pascal VOC, we use a learning rate of 0.001 for 80k iterations and 0.0001 for 30k iterations. For training on the other datasets, we use a learning rate of 0.001 for 50k iterations and 0.0001 for 20k iterations. We trained our model in the end-to-end manner with Stochastic Gradient Descent (SGD), where the momentum is 0.9, and the weight decay is 0.0005. Our program is implemented by Tensorflow \cite{abadi2015tensorflow:} on a GPU of GeForce GTX TITAN X.

\subsection{Overall Performance}

\paragraph{Performance on Pascal VOC benchmark}We compare our approach with several state-of-the-arts in this subsection. Results in terms of mean average precision (mAP) are shown in Table~\ref{table1}. Our model achieves the second best performance among all methods, which is 1.2\% lower than that of RON \cite{kong2017ron:} but 3.2\% higher than that of baseline Faster R-CNN. Besides, it is happy to see that our method outperforms ION \cite{bell2016inside-outside} with the same backbone network which used features from Conv3\_3, Conv4\_3 and Conv5\_3 to leverage context and multi-scale knowledge for object detection. From the table, we can see that although C-RPNs is designed aiming to improve detection in real world with imbalance data, it gets competitive performance on Pascal VOC benchmark.

\begin{table*}[htbp]
\centering
\caption{ Results on PASCAL VOC 2007 test set. 07+12: union of Pascal VOC07 trainval and VOC12 trainval.}
\label{table1}
\resizebox{\textwidth}{12mm}{
\begin{tabular}{lllllllllllllllllllllll}
\hline
Method & Trainset & mAP & aero & bike & bird & boat & bottle & bus & car & cat & chair & cow & table & dog & horse & mbike & person & plant & sheep & sofa& train& tv  \\ \hline
Fast R-CNN \cite{girshick2015fast} & 07+12 & 70.0 & 77.0 & 78.1 & 69.3 & 59.4 & 38.3 & 81.6 & 78.6 & 86.7 & 42.8 & 78.8 & 68.9 & 84.7 & 82.0 & 76.6 & 69.9 & 31.8 & 70.1 & 74.8 & 80.4 & 70.4 \\
Faster R-CNN \cite{ren2017faster} & 07+12 & 73.2 & 76.5 & 79.0 & 70.9 & 65.5 & 52.1 & 83.1 & 84.7 & 86.4 & 52.0 & 81.9 & 65.7 & 84.8 & 84.6 & 77.5 & 76.7 & 38.8 & 73.6 & 73.9 & 83.0 & 72.6 \\
SSD500 \cite{liu2016ssd:} & 07+12 & 75.1 & 79.8 & 79.5 & 74.5 & 63.4 & 51.9 & 84.9 & 85.6 & 87.2 & 56.6 & 80.1 & 70.0 & 85.4 & 84.9 & 80.9 & 78.2 & 49.0 & \textbf{78.4} & 72.4 & 84.6 & 75.5 \\
ION \cite{bell2016inside-outside} & 07+12 & 75.6 & 79.2 & \textbf{83.1} & \textbf{77.6} & 65.6 & 54.9 & 85.4 & 85.1 & 87.0 & 54.4 & 80.6 & \textbf{73.8} & 85.3 & \textbf{82.2} & 82.2 & 74.4 & 47.1 & 75.8 & 72.7 & 84.2 & \textbf{80.4} \\
RON \cite{kong2017ron:} & 07+12 & \textbf{77.6} & \textbf{86.0} & 82.5 & 76.9 & \textbf{69.1} & 59.2 & \textbf{86.2} & 85.5 & 87.2 & 59.9 & 81.4 & 73.3 & 85.9 & \textbf{86.8} & \textbf{82.2} & \textbf{79.6} & \textbf{52.4} & 78.2 & \textbf{76.0} & \textbf{86.2} & 78.0 \\
SIN \cite{liu2018structure} & 07+12 & 76.0 & 77.5 & 80.1 & 75.0 & 67.1 & 62.2 & 83.2 & 86.9 & \textbf{88.6} & 57.7 & \textbf{84.5} & 70.5 & \textbf{86.6} & 85.6 & 77.7 & 78.3 & 46.6 & 77.6 & 74.7 & 82.3 & 77.1 \\
\textbf{C-RPNs (ours)} & 07+12 & \underline{76.4} & 78.6 & 79.5 & 76.3 & 66.5 & \textbf{63.2} & 84.6 & \textbf{87.8} & 87.8 & \textbf{60.2} & 83.3 & 71.7 & 85.5 & 86.1 & 81.4 & 79.2 & 49.2 & 75.2 & 73.9 & 83.1 & 75.7 \\  \hline
\end{tabular}}
\end{table*}

\paragraph{Performance on BSBDV 2017} Table~\ref{table2} shows the comparisons of C-RPNs with state-of-the-arts on BSBDV 2017. As shown in Table~\ref{table2}, our method achieves the best performance and its average precision (AP) is 3.4\% higher than the second best (FPN \cite{lin2017feature}). More specifically, the AP of C-RPNs is 70.3\%, which obtains 11\% performance gain compared with that of Faster R-CNN. It is noted that our C-RPNs gets slightly lower mAP than that of RON \cite{kong2017ron:} on VOC 2007, but it outperforms RON by a margin of 12.3\% on BSBDV 2017. Also, the AP of C-RPNs is 8.8\% and 3.4\% higher than that of R-FCN \cite{dai2016r-fcn:} and FPN \cite{lin2017feature} respectively. These results demonstrate that our C-RPNs is more competitive in object detection in real world.

\begin{table*}[htbp]
\centering
\caption{Performance Comparison on BSBDV 2017.}
\label{table2}
\begin{tabular}{lll}
\hline
Method & Backbone Network & AP(\%)  \\ \hline
SSD500 \cite{liu2016ssd:} & VGG16 reduce & 42.0  \\
Faster R-CNN \cite{ren2017faster} & VGG16 & 59.3  \\
RON \cite{kong2017ron:} & ResNet-101 & 58.0  \\
R-FCN \cite{dai2016r-fcn:} & ResNet-50 & 61.5  \\
FPN \cite{lin2017feature} & ResNet-50 & \underline{66.9}  \\
SIN \cite{liu2018structure} & VGG16 & 58.4  \\
\textbf{C-RPNs (ours)} & VGG16 & \textbf{70.3}  \\  \hline
\end{tabular}
\end{table*}

\paragraph{Comparison with baseline Faster R-CNN on pedestrian datasets} Pedestrian datasets like Caltech pedestrian benchmark \cite{dollar2009pedestrian} and CityPersons \cite{zhang2017citypersons:} are more challenging then Pascal VOC, which are collected via monitoring cameras on realistic street scenes. Performances on these two datasets are helpful to verify the efficiency of our approach since the scales and occlusion of pedestrians are changed frequently. Table~\ref{table3} shows the comparisons of our C-RPNs with the baseline Faster R-CNN on these pedestrian datasets. Our C-RPNs achieves average precision of 48.1\% and 51.4\% on Caltech pedestrian benchmark and CityPersons, bringing 4.1\% and 2.3\% performance gain upon baseline Faster R-CNN, respectively, which indicates its robustness in intricate realistic scenes.

\begin{table*}[htbp]
\centering
\caption{Performance Comparison on Caltech Pedestrian Benchmark and CityPersons.}
\label{table3}
\begin{tabular}{lll}
\hline
Method & Caltech pedestrian benchmark & CityPersons  \\ \hline
Faster R-CNN \cite{ren2017faster} & 44.0 & 49.1  \\
\textbf{C-RPNs (ours)} & \textbf{48.1} & \textbf{51.4}  \\  \hline
\end{tabular}
\end{table*}

\subsection{Quantitive Examples}

\begin{figure}
	\centering
		\includegraphics[scale=.7]{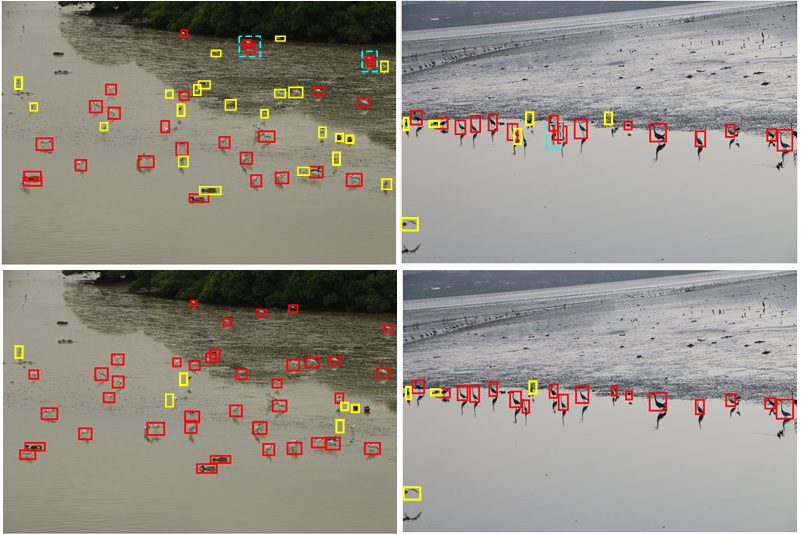}
	\caption{Detection results of Faster R-CNN (row 1) and our proposed C-RPNs (row 2) on BSBDV 2017. Best viewed in color.}
	\label{FIG:9}
\end{figure}

\paragraph{Qualitative Examples on wild bird detection}For visualization purpose, several examples of detection results on BSBDV 2017 are given in Figure~\ref{FIG:9}. The rows from the top to the bottom are respectively expressed as the results of Faster R-CNN and C-RPNs. Detection boxes from detectors are marked red. For better observation, we marked boxes of miss detection cases in yellow. According to the ground truth, there are 46 and 22 birds in the top and bottom images, respectively. From the results, we can see that our method shows significantly improved recall for object detection in wild scenes, where 40 and 17 birds have been detected, respectively. Compared with the results detected with Faster R-CNN, our method brings 16 and 2 more birds detected in two images respectively. Meanwhile, dotted boxes in blue show samples are detected with more than one boxes, three in the top images and none in the bottom images, which indicates that our method is able to generate more precise bounding boxes.

\begin{figure}
	\centering
		\includegraphics[scale=.29]{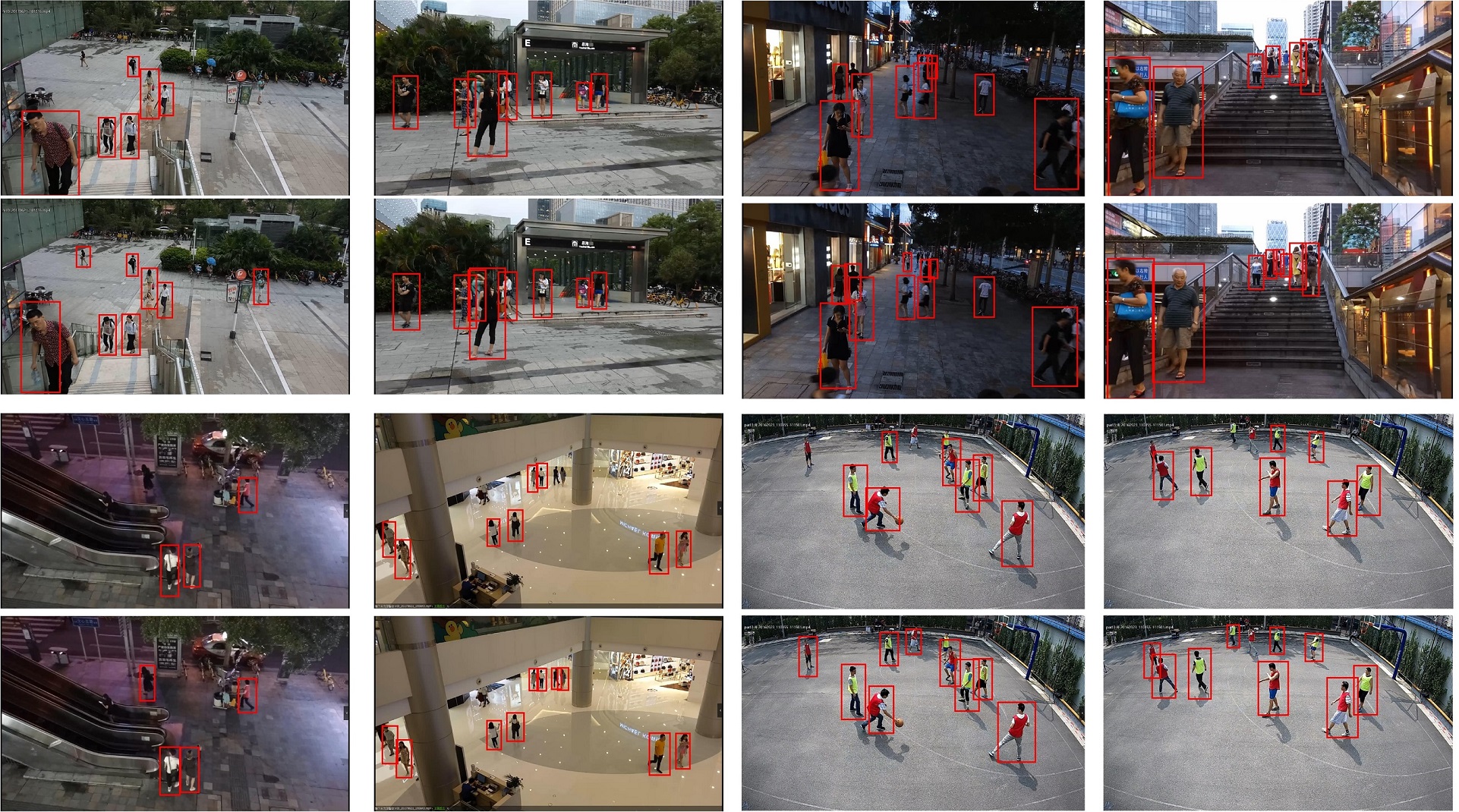}
	\caption{Detection results of Faster R-CNN (row 1 and row 3) and our proposed C-RPNs (row 2 and row 4) on realistic pedestrian images.}
	\label{FIG:10}
\end{figure}

\paragraph{Qualitative Examples on practical pedestrian detection}In Figure~\ref{FIG:10}, our proposed approach is trained on Caltech Pedestrian Benchmark and tested in realistic environments with random pedestrian flows. We show some detection images with different shooting angles such as looking down and looking up or with poor illumination, which are collected in subway, park and campus. Compared with the results from Faster R-CNN, our method brings more true positive and less false positive detections in these images respectively. It is found that some hard samples are falsely detected as background from Faster R-CNN, while those are detected aright as pedestrians from C-RPNs. According to the results in Figure~\ref{FIG:10}, our proposed C-RPNs can adapt to harsh and complex environments to provide high quality object detection in real world.

\subsection{Improvement analysis on false detections}

\begin{figure}
	\centering
		\includegraphics[scale=.32]{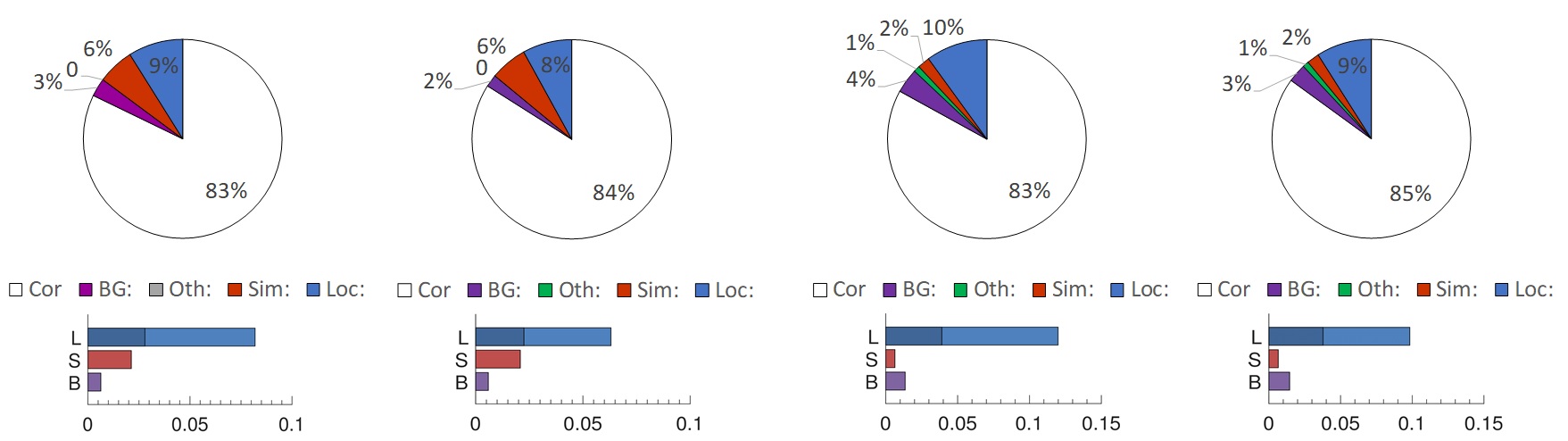}
	\caption{Analysis of Top-Ranked False Positives. Pie charts: fraction of detections that are correct (Cor) or false positive due to poor localization (Loc), confusion with similar objects (Sim), confusion with other VOC objects (Oth), or confusion with background (BG). Bar graphs: absolute AP improvement by removing all false positives of one type. ‘L’: first bar segment is improvement if duplicate or poor localizations are removed; second bar is improvement if localization errors are corrected so that the false positives become true positives. ‘B’: no confusion with background and non-similar objects. ‘S’: no confusion with similar objects. The first and second columns: results of the baseline Faster R-CNN and C-RPNs on detecting animals. The third and fourth columns: results of the baseline Faster R-CNN and C-RPNs on detecting vehicles.}
	\label{FIG:5}
\end{figure}

To further examine the improvement of our C-RPNs upon baseline Faster R-CNN, the analysis tools \cite{hoiem2012diagnosing} upon Pascal VOC are employed to produce a detailed error analysis. In Pascal VOC, animal categories include ‘bird’, ‘cat’, ‘cow’, ‘dog’, ‘horse’, ‘sheep’ and ‘person’. ‘Plane’, ‘bicycle’, ‘boat’, ‘bus’, ‘car’, ‘motorbike’ and ‘train’ make up the vehicle categories. Figure~\ref{FIG:5} takes animals and vehicles as examples to show the frequency and impact on the performance of each type of false positive. As shown, C-RPNs reduces detection errors compared with Faster R-CNN when detecting both animals and vehicles. It is found that C-RPNs has less BG errors as well as Loc errors compared with the baseline, indicating that C-RPNs can classify and localize objects better because it mined hard samples during training and learned stronger classifiers. However, just like Faster R-CNN, detection results from C-RPNs have same confusions with similar object categories, partly because binary classifiers in cascade RPN module only indicate samples to be background or object, which has limited promotion on distinguishing the categories of an object.

\begin{figure}
	\centering
		\includegraphics[scale=.25]{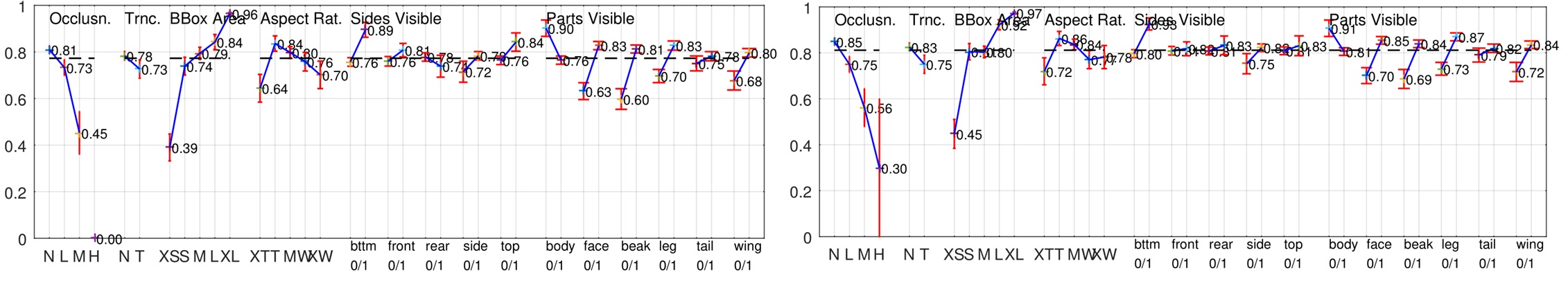}
	\caption{Characteristics analysis of different bird characteristics on VOC2007 test set: Each plot shows Normalized AP (APN \cite{hoiem2012diagnosing}) with standard error bars (red). Black dashed lines indicate overall APN. Key: Occlusion: N=none; L=low; M=medium; H=high. Truncation: N=not truncated; T=truncated Bounding Box Area: XS=extra-small; S=small; M=medium; L=large; XL =extra-large. Aspect Ratio: XT=extra-tall/narrow; T=tall; M=medium; W=wide; XW =extra-wide. Viewpoint / Part Visibility: ’1’=part/side is visible; ’0’=part/side is not visible. Left: results of the baseline Faster R-CNN. Right: results of C-RPNs.}
	\label{FIG:6}
\end{figure}

\begin{figure}
	\centering
		\includegraphics[scale=.27]{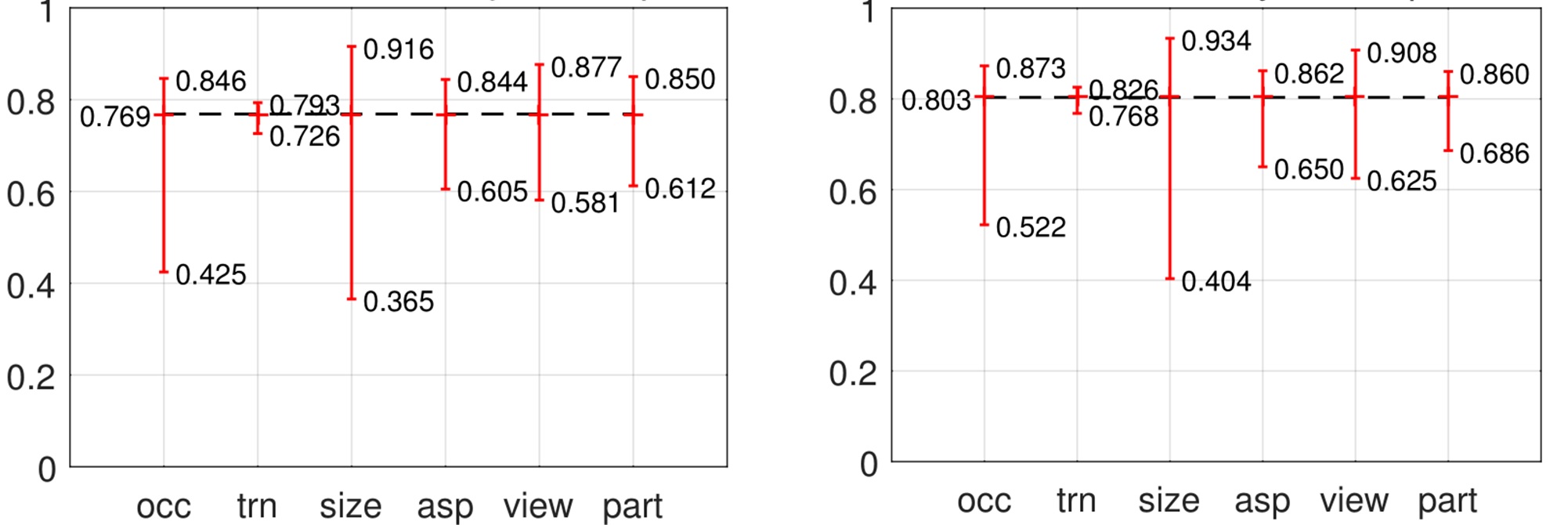}
	\caption{Summary of Sensitivity and Impact of Object Characteristics. The APN are shown over 7 categories of the highest performing and lowest performing subsets within each characteristic (occlusion, truncation, bounding box area, aspect ratio, viewpoint, part visibility). Overall APN is indicated by the dashed line. The difference between max and min indicates sensitivity. The difference between max and overall indicates the impact. Left: results of the baseline Faster R-CNN. Right: results of C-RPNs.}
	\label{FIG:7}
\end{figure}

Figure~\ref{FIG:6} visualizes the analysis of different bird characteristics on VOC2007 test set. Performance improvements on these characteristics are explicit since C-RPNs achieves higher average precision on all characteristics of Occlusion, Truncation, Bounding Box Area, Aspect Ratio, Viewpoint and Part Visibility. It is worth mentioning that when occlusion is High, C-RPNs can still recognize some birds while baseline Faster R-CNN detects nothing. Furthermore, extra annotations of seven categories(‘airplane’, ‘bicycle’, ‘bird’, ‘boat’, ‘cat’, ‘chair’, ‘table’) are created in \cite{hoiem2012diagnosing} for evaluating robustness of detection approaches. Figure~\ref{FIG:7} provides a compact summary of the sensitivity to each characteristic and the potential impact of improving robustness on seven categories. Overall, our C-RPNs achieves higher normalize average precision than Faster R-CNN against all characteristics, indicating its robustness in various scenes. Moreover, sensitivity against all these characteristics are decreased, which verifies that C-RPNs realizes an all-sided improvements upon Faster R-CNN. On the other side, we can see that C-RPNs is sensitive to the bounding box size just like Faster R-CNN and there is still some room to improve.

\subsection{Ablation Studies}

In previous sections, we have shown the efficiency of C-RPNs on several datasets. To further evaluate the individual effect of components of our C-RPNs, we analyze the object detection performance affected by the cascade stages as well as feature chain and score chain. We use BSBDV 2017 in this section.

\paragraph{Effects of cascade stages}To learn the efficiency of our C-RPNs with different number of cascade stages, results are summarized in Table~\ref{table4}. We remove different stages of C-RPNs to demonstrate their individual effect. It can be seen that, with stage \emph{3} and stage \emph{4}, C-RPNs achieves AP of 69.5\% which already outperforms the baseline Faster R-CNN. Adding stage \emph{2} and stage \emph{1} yields AP of 69.9\% and 70.3\% respectively, and it brings 0.4\% and 0.4\% performance gain respectively. Finally, the 4-stage cascade RPNs achieves the best performance. These results validate that employing more cascade stages and classifiers in the C-RPNs benefits the detection performance.

\begin{table*}[htbp]
\centering
\caption{The impact of cascade stages (BSBDV 2017).}
\label{table4}
\begin{tabular}{llll}
\hline
AP of C-RPNs(\%) & 69.5 & 69.9 & 70.3  \\ \hline
C-RPNs with Stage \emph{4} & \checkmark & \checkmark  & \checkmark \\
C-RPNs with Stage \emph{3} & \checkmark & \checkmark  & \checkmark \\
C-RPNs with Stage \emph{2} &            & \checkmark  & \checkmark \\
C-RPNs with Stage \emph{1} &            &             & \checkmark \\ \hline
\end{tabular}
\end{table*}

\paragraph{Effects of feature chain and score chain}To learn the impact of feature chain and score chain more specifically, Table~\ref{table5} shows the results of our C-RPNs with or without feature chain and score chain. We set the same parameters for C-RPNs with previous sections but control the usage of feature chain and score chain separately. As shown in Table~\ref{table5}, feature chain is found to be effective in C-RPNs, which brings 0.6\% performance gain. When we adapt score chain but without feature chain, the AP is 0.4\% higher, which illustrates the efficiency of using score chain as well. The adjustment boosts the performance by 0.9\% while both feature chain and score chain are used. These results verify that using features and scores at previous stages as the prior knowledge for the latter stages promotes the final detection.

\begin{table*}[htbp]
\centering
\caption{The impact of feature/score chain (BSBDV 2017).}
\label{table5}
\begin{tabular}{lllll}
\hline
AP of C-RPNs(\%) &69.4 & 70.0 & 69.8 & 70.3 \\ \hline
Feature Chain &        & \checkmark  &            & \checkmark  \\
Score Chain &          &             & \checkmark & \checkmark \\  \hline
\end{tabular}
\end{table*}

\begin{figure}
	\centering
		\includegraphics[scale=.32]{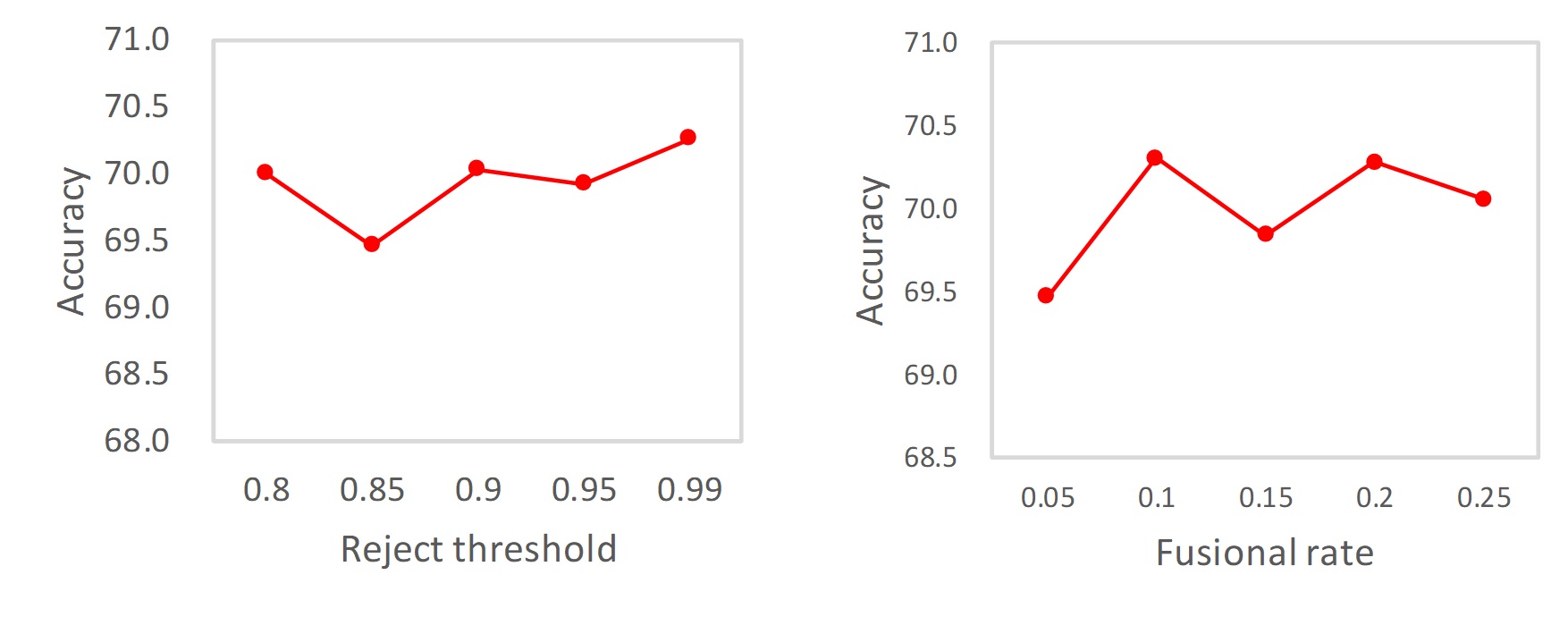}
	\caption{Grid search for the best reject threshold and fusion rate. Left: accuracy vs reject threshold \emph{r}; Right: accuracy vs fusion rate \emph{$\lambda_f$}.}
	\label{FIG:8}
\end{figure}

\paragraph{Selection of reject threshold and fusion rate}To find the best hyper parameters, empirical tests were conducted using different reject threshold \emph{r} and fusion rate \emph{$\lambda_f$} on BSBDV 2017 through one-dimensional grid search. Figure~\ref{FIG:8} shows the impacts of these two factors. As shown, reject threshold \emph{r=0.99} achieved the best AP of 70.31\% when the fusion rate was fixed at \emph{0.1}. We then fixed the reject threshold as \emph{0.99} and applied a grid search by changing the fusion rate \emph{$\lambda_f$}. From Figure~\ref{FIG:8}, the best \emph{$\lambda_f$} is observed as  \emph{0.1} with the AP of 70.31\%.

\section{Conclusion}

In this paper, we have constructed C-RPNs, an effective approach for object detection in real world. The essence of our C-RPNs lies in adopting a cascade structure of region proposal networks to discard easy samples during training and learn stronger classifiers. Moreover, a feature chain and a score chain at multiple stages have been proposed to help generating more discriminative representations for proposals. Finally, a loss function of cascade stages is designed to jointly learn cascade classifiers. Extensive experiments have been conducted to evaluate our C-RPNs on a common benchmark (Pascal VOC) and several challenging datasets collected in wild scenes or realistic traffic scenes. Our C-RPNs achieves competitive results compared with the current state-of-the-arts and outperforms the baseline Faster R-CNN by an obvious margin, demonstrating its efficacy for object detection in real world.

\bibliography{myReference}

\end{document}